\documentclass[letterpaper]{article} 
\usepackage{aaai2026}
\usepackage{times}  
\usepackage{helvet}  
\usepackage{courier}  
\usepackage[hyphens]{url}  
\usepackage{graphicx} 
\usepackage{natbib}  
\usepackage{caption} 
\usepackage{amsmath}
\usepackage{algorithm}
\usepackage{algorithmic}

\usepackage{longtable}
\usepackage{booktabs}

\usepackage{booktabs}
\usepackage{multirow}
\usepackage{amssymb}
\usepackage{xcolor}
\usepackage{colortbl}
\usepackage[breakable]{tcolorbox}
\usepackage{enumitem}
\usepackage{xurl}

\usepackage{newfloat}
\usepackage{listings}
\DeclareCaptionStyle{ruled}{labelfont=normalfont,labelsep=colon,strut=off} 
\floatstyle{ruled}
\newfloat{listing}{tb}{lst}{}
\floatname{listing}{Listing}
\title{CDPR: Counterfactual Advantage-based Credit Assignment for Cost-Aware Sequential Medical Diagnosis}

\newcommand{\corrauth}{\textsuperscript{\ensuremath{\dagger}}}

\author{
Qi Peng\textsuperscript{\rm 1,2,3},
Yi Cai\textsuperscript{\rm 1,2}\corrauth,
Changmeng Zheng\textsuperscript{\rm 3}\corrauth,
Xin Wu\textsuperscript{\rm 1,2},
Jiayuan Xie\textsuperscript{\rm 3},
Qing Li\textsuperscript{\rm 3}
}

\affiliations{
\textsuperscript{\rm 1}South China University of Technology, Guangzhou, China\\
\textsuperscript{\rm 2}Key Laboratory of Big Data and Intelligent Robot (SCUT), Ministry of Education, Guangzhou, China\\
\textsuperscript{\rm 3}The Hong Kong Polytechnic University, Hong Kong SAR, China\\
\textsuperscript{\ensuremath{\dagger}}Corresponding authors.
}

\usepackage{bibentry}

\begin{document}

\maketitle

\begin{abstract}

Clinical diagnosis is a step-by-step, cost-aware process: a physician orders examinations one at a time, observes the results, and updates the diagnosis before reaching a final conclusion. Most medical language models instead treat diagnosis as a one-pass classification task and ignore the trade-off between a test's value and its cost. We model diagnosis as a cost-aware sequential decision process and train the policy with reinforcement learning. 
The main difficulty is credit assignment: the only reliable signal comes once at the end of a long trajectory, so it scores a wasteful workup the same as an efficient one. 
We propose CDPR (Counterfactual Diagnostic Process Reward), which needs no expert labels and no learned critic. CDPR first finds the states where the policy hesitates, using the uncertainty of its action distribution, and then scores the chosen action by its advantage over the alternatives the policy itself would consider, estimated with short rollouts under a utility that balances correctness against test count, cost, and infeasible requests. 
A rollout cache reuses within-batch trajectories to keep the cost low. 
We integrate CDPR into GRPO and test it on one in-domain (MIMIC-IV) and two out-of-domain (ClinicalBench and a private hospital dataset) benchmarks. 
CDPR improves diagnostic accuracy while clearly reducing the number and cost of examinations.
Source code is available at \url{https://github.com/pqpq17/CDPR/tree/main}.
\end{abstract}


\section{Introduction}
\begin{figure}[!t]
\centering
\includegraphics[width=\linewidth]{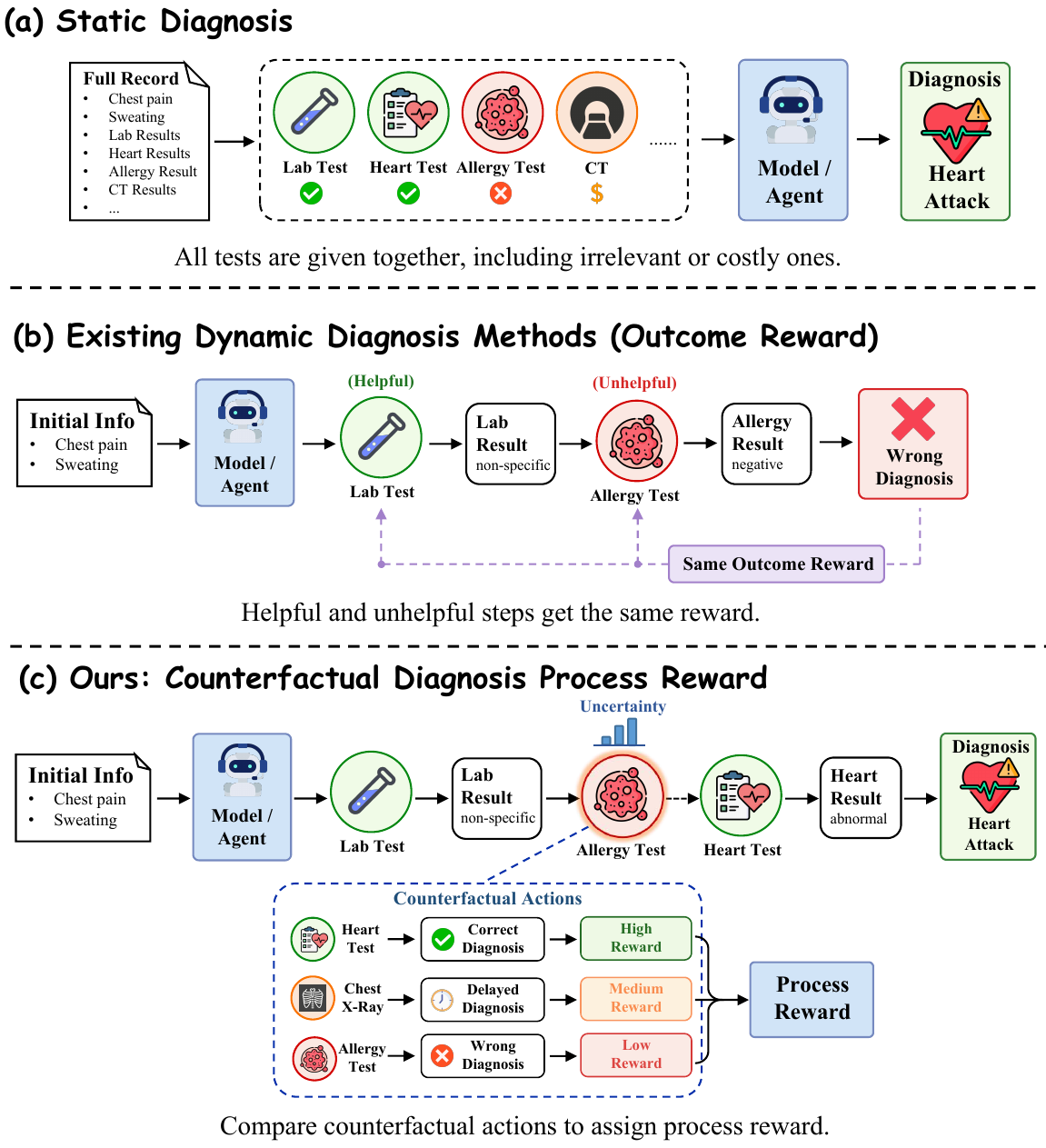}
\caption{Comparison of three diagnostic paradigms. (a) \emph{Static diagnosis}: the model is fed the full record at once, including irrelevant or costly examinations, and is asked to output the diagnosis in a single pass. (b) \emph{Existing dynamic diagnosis methods}: the model acquires evidence step by step but receives the same outcome reward for every intermediate action. (c) \emph{Ours:} We contrast the chosen action with counterfactual alternatives to establish process reward, providing dense supervision that rewards informative actions and penalizes redundant or low-value ones.}
\label{fig:intro}
\end{figure}

Clinical diagnosis is not a one-shot decision but an evolving, cost-aware process~\cite{ball2015improving,qiu2025evolving,nori2025sequential}. Physicians rarely commit to a conclusion upon first contact with a patient; instead, they begin from initial symptoms, weigh which examination would most reduce uncertainty, order it, observe the outcome, revise  differential, and only commit to a final diagnosis once the accumulated evidence is sufficient.
Most existing medical language models and diagnostic agents, however, sidestep this dynamic entirely: they treat diagnosis as a static classification problem in which the complete case description is consumed in a single pass and the final answer is produced directly~\cite{jin2021disease,singhal2023large}. A typical example is shown in Figure \ref{fig:intro} (a) where the final diagnosis (\textit{i.e., heart attack}) is delivered with all possible test results, even including irrelevant or very costly ones. Such a formulation captures \textit{what} the correct diagnosis is, but ignores \textit{how} one should efficiently arrive at it.
A competent clinician may implicitly balance the marginal diagnostic value of an additional test against its price~\cite{pauker1980threshold}.

To recover this missing dimension, it is natural to model diagnosis as a sequential decision process and to train the agent with reinforcement learning, rewarding it for reaching the correct diagnosis through its own sequence of examination and diagnosis actions.
Yet this formulation immediately surfaces a limitation: the only reliable supervision signal, whether the final diagnosis is correct, arrives once, at the very end of a long trajectory. 
This sparse, outcome-only reward gives rise to a classic \textit{credit assignment} problem~\cite{pignatelli2023survey,lightman2024let}.
Two trajectories that reach the same correct diagnosis may differ dramatically along the way: one may resolve the case with a handful of targeted tests, while another orders redundant or needlessly expensive examinations before stumbling onto the same answer.
The outcome reward scores these two trajectories identically (as shown in Figure \ref{fig:intro} (b)), offering no signal that steers the policy toward shorter, cheaper, and more clinically sound workups.
\textbf{Attributing the trajectory-level outcome to the intermediate actions that genuinely drive it is the central challenge we address.}

The standard solutions for sparse rewards, however, fit the medical setting poorly. 
One line of work trains process reward models, or auxiliary critic functions, to provide denser feedback for intermediate decisions \cite{lightman2024let,uesato2022solving}, but this presupposes dense intermediate supervision that is unattainable here: clinicians qualified to label diagnostic trajectories are scarce and costly, and the supervision is intrinsically ill-defined, since a correct diagnosis can legitimately be reached through many different examination paths and there is rarely a single "ground-truth" next action experts would agree upon. 
An alternative is to estimate the value of intermediate actions by Monte-Carlo rollout \cite{yao2023tree,wang2024math,kazemnejad2024vineppo}, but performing such rollouts at every intermediate state is prohibitively expensive and largely wasteful. 
Neither annotation-heavy critics nor brute-force rollouts offer a viable path to dense supervision in clinical diagnosis.

These difficulties point to two observations that organize our approach. 
First, not every state deserves evaluation. Only the decision-critical states at which the policy genuinely hesitates, where the current evidence is insufficient to commit to a single next action and alternative choices would lead to materially different downstream trajectories. 
Second, value should be measured counterfactually and relatively: whether an action is good is not an absolute property but a comparison against the alternatives the policy itself would plausibly consider at that state.
We formalize this as \textit{counterfactual advantage-based credit assignment}, as shown in Figure \ref{fig:intro} (c).
On each selected state, we estimate the action's value through short-horizon Monte-Carlo rollouts under a utility that balances diagnostic correctness against examination count, cost, and infeasible requests, and we measure its advantage over a baseline formed by the policy's own top-ranked counterfactual actions.
To make this estimation affordable, we introduce a rollout cache that reuses trajectories already generated within a training batch, reducing variance and cost while, as a by-product, surfacing a second signal of decision instability. 
The resulting process reward is dense, annotation-free, and critic-free, and it redistributes the sparse outcome signal onto the intermediate actions that actually drive the result.

We instantiate this idea as \textbf{CDPR (Counterfactual Diagnostic Process Reward)} and integrate it seamlessly into Group Relative Policy Optimization, layering a state-level counterfactual advantage onto GRPO's trajectory-level group-normalized advantage to form a hierarchical credit-assignment scheme. 
Our contributions are summarized as follows:

\begin{itemize}
    \item We reformulate medical diagnosis as a cost-aware sequential decision process and identify outcome-only credit assignment as its core obstacle.
    \item We propose CDPR, which casts process-reward estimation as counterfactual advantage estimation, concentrating an uncertainty-guided budget on decision-critical states and reusing prior rollouts through a cache to make estimation efficient.
    \item We integrate CDPR into GRPO and conduct extensive experiments on one in-domain (MIMIC-IV) and two out-of-domain (ClinicalBench and a private hospital dataset) benchmarks, showing that CDPR consistently improves diagnostic accuracy while substantially reducing the number and cost of examinations.
\end{itemize}

\section{Related Work}

With the rapid development of artificial intelligence, an increasing number of models have been applied to medical tasks~\cite{peng2025aligning}, including medical question answering~\cite{peng2024integration}, medical report generation~\cite{jin2024promptmrg,yan2022clinical}, and medical image analysis~\cite{chen2019synergistic,wu2024voco,peng2025cke,liang2025seeing}. These tasks are denoted as static diagnosis: the model is presented with a complete case description in a single shot and is asked to produce a diagnostic conclusion in one pass. 
This static formulation ignores the sequential nature of real clinical practice, where evidence is collected over time. It also fails to support cost-aware decisions, where the diagnostic value of an examination must be balanced against its cost and time burden.

\begin{figure*}[t]
  \centering
  \includegraphics[width=\linewidth]{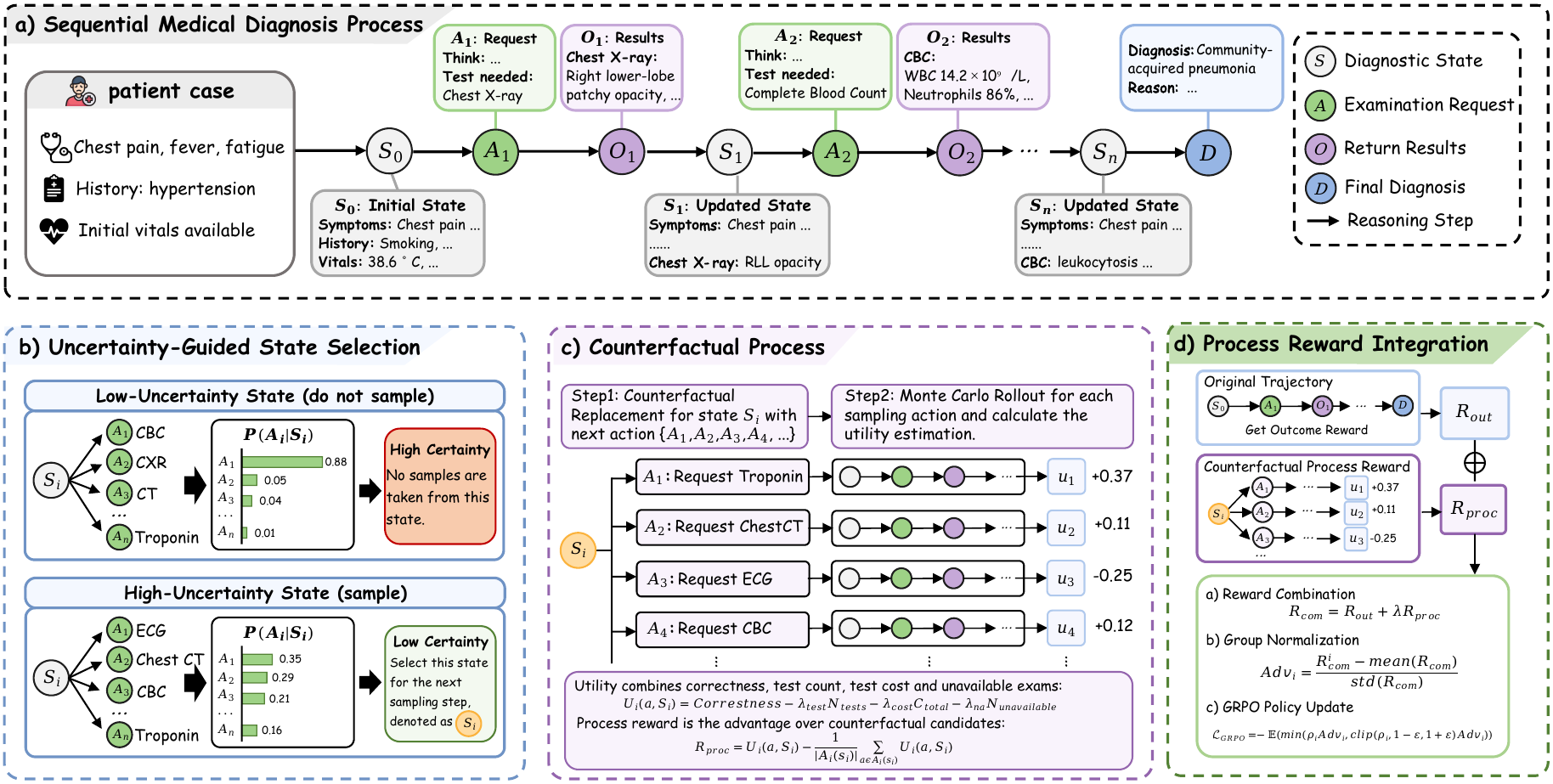}
      \caption{The overview of the framework. (a) We reformulate medical diagnosis as a sequential decision process in which the agent interleaves examination and diagnosis actions. (b) For each trajectory, we identify decision-critical states using the action uncertainty of the current actor. (c) On the selected states, we estimate a counterfactual process reward by rolling out alternative actions and reusing prior trajectories through a rollout cache. (d) The process reward is integrated with the outcome reward to train the policy under the GRPO objective.}
  \label{fig:framework}
\end{figure*}

Regardless of how the diagnostic process is structured, the model must ultimately reason over evidence \cite{liang2026multi,zheng2025learning} grounded in medical knowledge, an ability rooted in advances on knowledge extraction \cite{zheng2023rethinking} and reasoning \cite{zheng2026multimodal}. Based on this, recent work has increasingly modeled diagnosis as a sequential decision process. In this setting, evidence is collected step by step, reflecting the dynamic nature of real clinical practice. Multi-agent simulators approximate clinical practice by constructing clinical environments and patient simulators, where agents interact with patients and make diagnostic decisions~\cite{schmidgall2024agentclinic,almansoori2025medagentsim,tang2024medagents,kim2024mdagents,peng2025tree,nori2025sequential,he2026evoclinician,zheng2024picture}. Although these methods model diagnosis as a sequential decision process, the agents themselves are typically driven by prompting and rely heavily on backbone capability rather than learned diagnostic policies.
More recent works train diagnostic policies with reinforcement learning. For instance, Qiu \emph{et al.} train an agent in a virtual clinical environment with agentic reinforcement learning~\cite{qiu2025evolving}. These methods rely on sparse outcome rewards and therefore suffer from severe credit-assignment difficulty along long diagnostic trajectories. 
To handle this challenge, recent studies have explored step-wise or process-level rewards that assign credit to intermediate actions. For example, ReasonRAG~\cite{zhang2026process}, StepSearch~\cite{wang2025stepsearch}, and GiGPO~\cite{feng2026group} rely on densely sampling continuations or alternative actions from intermediate states to estimate step-level values without extra annotation. Although these methods provide denser supervision than outcome-only rewards, they may suffer from limited scalability as performing such rollouts at every state is expensive and wasteful.

\section{Methodology}

\subsection{Problem Formulation}
\label{sec:formulation}

Most existing work formulates medical diagnosis as a static classification task, where a model is given the complete case description in a single shot and directly produces the final diagnosis. In real clinical practice, however, diagnosis is inherently sequential: a physician starts from initial symptoms, iteratively orders examinations, observes their results, updates the differential, and only commits to a final diagnosis once the evidence is sufficient. To better capture this dynamic situation, we model medical diagnosis as a sequential decision process, as illustrated in Figure~\ref{fig:framework} (a).

\noindent\textbf{Sequential Diagnosis as an MDP.} We model an episode as a Markov Decision Process $\mathcal{M}=(\mathcal{S}, \mathcal{A}, \mathcal{T}, R)$. Each state
\begin{equation}
s_t = (x_0, \{(a_i, o_i)\}_{i=0}^{t-1})
\end{equation}
encodes the initial patient presentation $x_0$ together with the history of examinations issued so far and their observed results. At each step, the policy $\pi_\theta$ generates an action in free-form text and then parses it into one of two types,
\begin{equation}
a_t \in \mathcal{A} = \mathcal{A}^{\text{exam}} \cup \mathcal{A}^{\text{diag}},
\end{equation}
where $\mathcal{A}^{\text{exam}}$ denotes examination actions, each specifying an examination to be ordered, and $\mathcal{A}^{\text{diag}}$ denotes diagnosis actions, each specifying a final diagnosis to commit to. The action is sampled from a policy $\pi_\theta$ generated by the underlying language model, $a_t \sim \pi_\theta(\cdot \mid s_t)$, rather than chosen from a fixed candidate list. Issuing an examination $a_t \in \mathcal{A}^{\text{exam}}$ deterministically extends the state with the resulting evidence, $s_{t+1} = s_t \oplus (a_t, o_t)$, while emitting a diagnosis action $a_t \in \mathcal{A}^{\text{diag}}$ terminates the episode. A trajectory is therefore a sequence $\tau = (s_0, a_0, \ldots, s_T, a_T)$ with $a_T = \hat{d} \in \mathcal{A}^{\text{diag}}$. Given the ground-truth diagnosis $d^{\star}$, the outcome reward of $\tau$ is
\begin{equation}
R_{\text{out}}(\tau) = \mathbf{1}(\hat{d} = d^{\star}).
\label{eq:outcome}
\end{equation}

\noindent\textbf{Credit Assignment Challenge.} Training a policy $\pi_\theta$ purely with $R_{\text{out}}$ provides only a single sparse signal at the end of each trajectory and cannot resolve which intermediate actions actually drive the outcome. In particular, two trajectories that arrive at the same final diagnosis may differ substantially along the way: one may reach the answer with a few targeted examinations, while another issues many redundant tests or repeatedly picks expensive examinations when cheaper alternatives carry comparable diagnostic value. The outcome reward treats these trajectories identically, and therefore offers no signal that steers the policy toward shorter and more cost-effective workups. This motivates a process reward that scores intermediate actions according to their long-term diagnostic value, which we develop in the next three sections.

\subsection{Action Uncertainty-guided State Selection}
\label{sec:selection}

A natural way to provide such a process reward is to learn a critic that judges intermediate actions, but in the medical domain this is hardly feasible. Clinical experts qualified to label diagnostic trajectories are scarce and expensive, and the supervision itself is intrinsically ill-defined: a correct diagnosis can be reached through many legitimate examination paths, so there is rarely a single ``ground-truth'' next action that experts can agree on. We therefore aim at a process reward that requires no human annotation. Inspired by Monte Carlo value estimation, we estimate the long-term diagnostic value of an action at a state by sampling continuations from it under the current policy and scoring their final outcomes. Building on this idea, for each intermediate state we substitute the actually taken action with alternative \emph{counterfactual} actions, roll out short trajectories from each, and assign the actually taken action a reward equal to its advantage in expected diagnostic outcome over these alternatives.

Performing such counterfactual rollouts at every intermediate state, however, is prohibitively expensive and largely wasteful, since most intermediate states are routine and the policy's decision is essentially determined by the evidence already collected. The states that genuinely deserve counterfactual evaluation are those at which the policy \emph{hesitates}, i.e., states where the current evidence is insufficient to commit to a single next action and where alternative choices may lead to substantially different downstream trajectories. We therefore concentrate the counterfactual budget on such decision-critical states and identify them from the action distribution of the current actor, as illustrated in Figure~\ref{fig:framework}(b).

\noindent\textbf{Next-action Sampling.} For each intermediate state $s_t$ along a sampled trajectory, we draw $n_s$ next actions from the current policy:
\begin{equation}
\{\tilde{a}^{(1)}, \tilde{a}^{(2)}, \ldots, \tilde{a}^{(n_s)}\} \sim \pi_\theta(\cdot \mid s_t).
\end{equation}
Compared with launching full rollouts, this is a lightweight one-step probe that reveals how concentrated the local action distribution is.

\noindent\textbf{Action Uncertainty Score.} Let $\hat{p}(a \mid s_t)$ denote the empirical frequency of action $a$ in the $n_s$ samples and $\tilde{\mathcal{A}}(s_t)$ the set of distinct sampled actions. We measure the dispersion of the local distribution by its empirical entropy
\begin{equation}
\mathcal{U}(s_t) = -\sum_{a \in \tilde{\mathcal{A}}(s_t)} \hat{p}(a \mid s_t) \log \hat{p}(a \mid s_t).
\label{eq:uncertainty}
\end{equation}
A small $\mathcal{U}(s_t)$ indicates that the policy is confident at $s_t$, so counterfactual evaluation would yield a marginal benefit; a large $\mathcal{U}(s_t)$ indicates genuine hesitation among multiple examinations or between continuing to examine and committing to a diagnosis, and is precisely where redistributing credit is most informative.

\noindent\textbf{Entropy-based Pre-selection.} Given an uncertainty threshold $\eta$, we collect the high-uncertainty states of trajectory $\tau$:
\begin{equation}
\mathcal{S}^{\text{ent}}(\tau) = \{\, s_t \in \tau \;:\; \mathcal{U}(s_t) \geq \eta \,\}.
\label{eq:ent-set}
\end{equation}
The remaining states carry low action-distribution uncertainty and are unlikely to yield informative counterfactual signals. The high-uncertainty set $\mathcal{S}^{\text{ent}}(\tau)$ will later be combined with a complementary cache-driven signal to form the final set of states to evaluate.

\subsection{Counterfactual Process Reward}
\label{sec:rollout}

This section illustrates the counterfactual process reward estimation procedure, as shown in Figure~\ref{fig:framework}(c). For an intermediate state, we first specify a set of counterfactual candidate actions, estimate the long-term diagnostic value of each candidate by Monte Carlo simulation, and obtain the process reward of the actually taken action by comparing it against these candidates. To reduce the sampling cost, we further introduce a rollout cache that reuses trajectories already generated during training and, as a by-product, provides an additional signal for state selection. We unify this signal with the entropy-based method to form the final set of states to evaluate. The remainder of this section details each step.

\noindent\textbf{Counterfactual Action Candidates.} For an intermediate state $s$, we sample next actions from the current policy and rank them by their empirical probability. The top-$M$ distinct actions form the counterfactual candidate set
\begin{equation}
\mathcal{A}_c(s) = \mathrm{Top}\text{-}M\big(\hat{p}(\cdot \mid s)\big),
\label{eq:cand-set}
\end{equation}
where $\hat{p}(\cdot \mid s)$ is estimated from the same samples used in Eq.~\eqref{eq:uncertainty}. 

\noindent\textbf{Rollout Cache.} Counterfactual evaluation is the dominant cost of the method, since each selected state requires one rollout per candidate action. To reduce this cost, we reuse the rollouts already produced for each case during training. The same case typically yields multiple sampled trajectories within a training batch. We align these trajectories by their diagnostic prefix and identify intermediate states that share both the case description and the observed examination history. Whenever multiple aligned trajectories take different next actions at such a state, the suffixes of these trajectories form a pool of cached continuations $\mathcal{T}^{\text{cache}}_{s,a}$ for each branching action $a$, and can be used directly as Monte Carlo samples without launching new rollouts.

The cache also provides a second signal of action uncertainty. A state at which aligned trajectories disagree on the next action is, by definition, a state where the policy hesitated under nearly identical evidence. The disagreement is more informative when the diverging suffixes lead to materially different downstream behavior, for example, one path arrives at the correct diagnosis while another does not, or the paths differ markedly in the number of examinations issued, their total cost, or the number of unavailable requests. We mark such a state as cache-disagreed and define
\begin{equation}
\mathcal{S}^{\text{cache}}(\tau) = \{\, s_t \in \tau \;:\; s_t \text{ is cache-disagreed} \,\}.
\label{eq:cache-set}
\end{equation}

\noindent\textbf{Unified State Selection.} Next-action sampling estimates uncertainty from the local action distribution of the current actor, while the rollout cache estimates it from realised action disagreement on existing trajectories. The two signals serve the same purpose of locating states whose next decision is unstable and whose counterfactual evaluation is most informative. We therefore take their union as the final set of states to evaluate, with cache-disagreed states preferred when the budget is limited:
\begin{equation}
\mathcal{S}^{\star}(\tau) = \mathcal{S}^{\text{cache}}(\tau) \cup \mathcal{S}^{\text{ent}}(\tau).
\label{eq:selected}
\end{equation}

\noindent\textbf{Short-horizon Continuation.} For each $s \in \mathcal{S}^{\star}(\tau)$ and each $a \in \mathcal{A}_c(s)$, we obtain $K$ continuations from $(s, a)$. When $|\mathcal{T}^{\text{cache}}_{s,a}| \geq K$, we draw the continuations directly from the cache; otherwise we sample fresh rollouts from the current policy. To keep the cost of process-reward estimation manageable, each rollout is bounded by horizon $H$:
\begin{equation}
\{\tau^{(k)}_{s,a}\}_{k=1}^{K} \sim \pi_\theta(\cdot \mid s, a), \quad |\tau^{(k)}_{s,a}| \leq H,
\label{eq:rollout}
\end{equation}
where each $\tau^{(k)}_{s,a}$ either reaches a final diagnosis within $H$ steps or is truncated and forced to emit a diagnosis based on the evidence seen so far.

\noindent\textbf{Utility Function.} For a continuation $\tau$, we score its diagnostic value with a utility that combines the outcome reward in Eq.~\eqref{eq:outcome} with three effort signals observable from the environment: the number of issued examinations $N_{\text{tests}}(\tau)$, their total monetary cost $C_{\text{total}}(\tau)$, and the number of unavailable examinations requested $N_{\text{na}}(\tau)$:
\begin{equation}
U(\tau) = \mathbf{1}(\hat{d} = d^{\star}) - \lambda_{\text{test}} N_{\text{tests}}(\tau) - \lambda_{\text{cost}} C_{\text{total}}(\tau) - \lambda_{\text{na}} N_{\text{na}}(\tau),
\label{eq:utility}
\end{equation}
where $\lambda_{\text{test}}, \lambda_{\text{cost}}, \lambda_{\text{na}} \geq 0$ trade off correctness against examination effort, monetary cost, and clinically infeasible requests. Averaging over the $K$ continuations gives the action-value estimate
\begin{equation}
\hat{U}(s, a) = \frac{1}{K} \sum_{k=1}^{K} U(\tau^{(k)}_{s,a}).
\label{eq:uhat}
\end{equation}

\noindent\textbf{Counterfactual Process Reward.} The process reward of the actually taken action $a^{\dagger}$ is its advantage over the average utility of the candidate set:
\begin{equation}
R_{\text{proc}}(s, a^{\dagger}) = \hat{U}(s, a^{\dagger}) - \frac{1}{|\mathcal{A}_c(s)|} \sum_{a' \in \mathcal{A}_c(s)} \hat{U}(s, a').
\label{eq:rproc}
\end{equation}
By construction, $R_{\text{proc}}(s, a^{\dagger})$ is positive only when $a^{\dagger}$ leads to more accurate, cheaper, or shorter completions than the alternatives the policy itself would consider plausible at $s$, and negative otherwise. The trajectory-level outcome signal is thereby redistributed onto the intermediate actions that actually drive the result, without recourse to a learned critic or expert annotation.

\subsection{Training with Counterfactual Process Reward}
\label{sec:training}

We optimize $\pi_\theta$ with Group Relative Policy Optimization (GRPO), a critic-free variant of PPO that, for each case, samples a group of $n$ trajectories and normalizes their returns within the group to obtain advantages. GRPO removes the value network and stabilizes training in long-horizon decision tasks, which makes it a natural fit for our sequential diagnosis setting. We integrate the counterfactual process reward in Eq.~\eqref{eq:rproc} into this objective so that the outcome reward provides a long-horizon diagnostic target while the process reward densifies supervision at decision-critical states, as illustrated in Figure~\ref{fig:framework}(d).

\noindent\textbf{Combined Per-step Reward.} Along trajectory $\tau$, we place the outcome reward at the terminal step and the process reward at every selected decision state:
\begin{equation}
r_t(\tau) = \mathbf{1}(t = T)\, R_{\text{out}}(\tau) + \beta\, \mathbf{1}\!\left(s_t \in \mathcal{S}^{\star}(\tau)\right) R_{\text{proc}}(s_t, a_t),
\label{eq:combined}
\end{equation}
where $\beta \geq 0$ controls the strength of the process reward and $T$ is the terminal step. Non-selected intermediate states do not receive process reward, and the outcome reward is delivered exactly once at the trajectory end. In our GRPO implementation, these token-level rewards are summed into a single trajectory score
\begin{equation}
G(\tau) = R_{\text{out}}(\tau) + \beta \sum_{s_t \in \mathcal{S}^{\star}(\tau)} R_{\text{proc}}(s_t, a_t),
\label{eq:traj-score}
\end{equation}
which is then group-normalized within GRPO to obtain advantages.

\noindent\textbf{Group-relative Advantage.} For each case, GRPO samples a group of $n$ trajectories $\{\tau^{(i)}\}_{i=1}^{n}$ and normalizes the returns within the group:
\begin{equation}
A^{(i)} = \frac{G(\tau^{(i)}) - \mu_G}{\sigma_G}, \quad \mu_G = \frac{1}{n}\sum_{j=1}^{n} G(\tau^{(j)}),
\end{equation}
with $\sigma_G$ the empirical standard deviation. The same group of trajectories is also fed into the rollout cache, so that group sampling, state selection, and utility estimation share computation.

\noindent\textbf{Policy Update.} We optimize $\pi_\theta$ with a clipped surrogate objective:
\begin{equation}
\mathcal{L}(\theta) = -\,\mathbb{E}\Big[\sum_{t} \min\big( \rho_t(\theta) A^{(i)},\, \mathrm{clip}(\rho_t(\theta), 1-\epsilon, 1+\epsilon) A^{(i)} \big) \Big],
\end{equation}
where $\rho_t(\theta) = \pi_\theta(a_t \mid s_t) / \pi_{\theta_{\text{old}}}(a_t \mid s_t)$ and $\epsilon$ is the clipping range. Compared with vanilla GRPO that relies solely on $R_{\text{out}}$, our objective inherits its stability while replacing the trajectory-level signal with a counterfactual-aware advantage that adaptively densifies on decision-critical states.

\section{Experiments}

\subsection{Experimental Setup}

\noindent\textbf{Datasets.}
The training corpus contains 23{,}377 cases from \textbf{MIMIC-IV}, a public critical-care database in which each case records the patient's history, current condition, ordered examinations, individual examination cost, and the final diagnosis. We hold out 750 additional MIMIC-IV cases as the in-domain test set. For out-of-domain evaluation, we use \textbf{ClinicalBench}~\cite{yan2026clinicallab}, an end-to-end multi-departmental clinical diagnostic benchmark of 1{,}500 real cases covering 24 departments and 150 diseases, and a \textbf{Private Dataset} of 952 inpatient cases collected from a partner hospital with full admission-to-discharge records (history, current condition, examinations, results, and diagnosis). The two out-of-domain sets come from clinical environments and data distributions distinct from training, so jointly the three sets allow a comprehensive evaluation under varying clinical settings.

\begin{table*}[!t]
\centering
\caption{Main results on MIMIC-IV test set and two out-of-domain benchmarks (ClinicalBench, Private). We report Diagnostic Accuracy (Acc., \%), Average Number of Examinations (AEN), and Average Exam Cost (AEC, USD). For open-source models, the best results are in \textbf{bold} and the second are \underline{underlined}. Closed-source LLMs are reported in \textcolor{gray!135}{gray} for reference.}
\label{tab:main_results}
\setlength\tabcolsep{3.6pt}
\fontsize{8.6pt}{10.4pt}\selectfont
\begin{tabular}{lccccccccc|ccc}
\toprule
\multirow{2}[2]{*}{\textbf{Method}}
  & \multicolumn{3}{c}{\textbf{MIMIC-IV (In-domain)}}
  & \multicolumn{3}{c}{\textbf{ClinicalBench (OOD)}}
  & \multicolumn{3}{c}{\textbf{Private (OOD)}}
  & \multicolumn{3}{c}{\textbf{Average}} \\
\cmidrule(lr){2-4} \cmidrule(lr){5-7} \cmidrule(lr){8-10} \cmidrule(lr){11-13}
 & Acc. & AEN & AEC
 & Acc. & AEN & AEC
 & Acc. & AEN & AEC
 & Acc. & AEN & AEC \\
\midrule
\rowcolor[HTML]{f0f0f0}
\multicolumn{13}{l}{\textit{\textbf{Closed-Source LLMs}}} \\
GPT-5.4              & \textcolor{gray!135}{42.93} & \textcolor{gray!135}{2.20} & \textcolor{gray!135}{62.80}  & \textcolor{gray!135}{56.00} & \textcolor{gray!135}{1.59} & \textcolor{gray!135}{18.46}  & \textcolor{gray!135}{36.97} & \textcolor{gray!135}{1.17} & \textcolor{gray!135}{67.39}  & \textcolor{gray!135}{45.30} & \textcolor{gray!135}{1.65} & \textcolor{gray!135}{49.55} \\
Claude-Sonnet-4.6    & \textcolor{gray!135}{38.93} & \textcolor{gray!135}{4.65} & \textcolor{gray!135}{84.13}  & \textcolor{gray!135}{54.27} & \textcolor{gray!135}{4.88} & \textcolor{gray!135}{44.05}  & \textcolor{gray!135}{32.98} & \textcolor{gray!135}{2.61} & \textcolor{gray!135}{74.57}  & \textcolor{gray!135}{42.06} & \textcolor{gray!135}{4.05} & \textcolor{gray!135}{67.58} \\
Gemini-3.1-Flash     & \textcolor{gray!135}{34.93} & \textcolor{gray!135}{2.37} & \textcolor{gray!135}{54.86}  & \textcolor{gray!135}{51.67} & \textcolor{gray!135}{1.90} & \textcolor{gray!135}{15.13}  & \textcolor{gray!135}{32.98} & \textcolor{gray!135}{1.78} & \textcolor{gray!135}{55.21}  & \textcolor{gray!135}{39.86} & \textcolor{gray!135}{2.02} & \textcolor{gray!135}{41.73} \\
DeepSeek-V4          & \textcolor{gray!135}{40.00} & \textcolor{gray!135}{3.76} & \textcolor{gray!135}{71.90}  & \textcolor{gray!135}{54.00} & \textcolor{gray!135}{2.69} & \textcolor{gray!135}{25.85}  & \textcolor{gray!135}{29.94} & \textcolor{gray!135}{2.66} & \textcolor{gray!135}{68.23}  & \textcolor{gray!135}{41.31} & \textcolor{gray!135}{3.04} & \textcolor{gray!135}{55.33} \\
\midrule
\rowcolor[HTML]{f0f0f0}
\multicolumn{13}{l}{\textit{\textbf{Open-Source LLMs}}} \\
Qwen3-4B \cite{yang2025qwen3}            & 32.40 & \underline{1.38} & \underline{39.80}  & 44.87 & \textbf{1.13} & 20.32  & 25.95 & \textbf{1.08} & 35.34  & 34.41 & \textbf{1.20} & \underline{31.82} \\
Qwen3-8B \cite{yang2025qwen3}            & 32.53 & 2.84 & 75.29  & 47.80 & 1.97 & 28.10  & 32.88 & 1.99 & 55.52  & 37.74 & 2.27 & 52.97 \\
Qwen3-32B \cite{yang2025qwen3}           & 37.73 & 3.12 & 97.81  & \underline{55.60} & 1.96 & 32.02  & \underline{33.40} & 2.01 & 60.20  & 42.24 & 2.36 & 63.34 \\
Ministral-3-8B \cite{liu2026ministral}      & 32.40 & 2.47 & 64.59  & 46.80 & \underline{1.66} & \underline{18.19}  & 19.64 & 1.90 & 36.17  & 32.95 & \underline{2.01} & 39.65 \\
Med-Gemma-27B \cite{sellergren2025medgemma}        & 35.60 & 3.35 & 106.41 & 54.80 & 1.80 & 28.59  & 32.25 & 1.92 & 59.68  & 40.88 & 2.36 & 64.89 \\
Huatuo-o1-8B \cite{chen2024huatuogpto1medicalcomplexreasoning}       & 12.40 & 8.43 & 108.80 & 28.53 & 7.58 & 74.95  & 16.28 & 7.40 & 82.26  & 19.07 & 7.80 & 88.67 \\
Baichuan-M1-32B \cite{wang2025baichuan}     & 19.20 & 8.49 & 81.96  & 36.40 & 6.20 & 28.30  & 16.50 & 6.30 & 58.20  & 24.03 & 7.00 & 56.15 \\
\midrule
\rowcolor[HTML]{f0f0f0}
\multicolumn{13}{l}{\textit{\textbf{Medical Agents}}} \\
MedAgent~\cite{tang2024medagents}    & 33.60 & 2.40 & 70.22 & 49.40 & 1.70 & 22.19 & 30.78 & 2.02 & 60.50 & 37.93 & 2.04 & 50.97 \\
MDAgent~\cite{kim2024mdagents}       & 34.53 & 2.65 & 76.88 & 49.93 & 1.93 & 24.30 & 30.36 & 2.29 & 65.92 & 38.27 & 2.29 & 55.70 \\
ToR~\cite{peng2025tree}              & 38.27 & 2.95 & 85.34 & 50.73 & 2.10 & 28.13 & 30.25 & 2.51 & 72.43 & 39.75 & 2.52 & 61.97 \\
\midrule
\rowcolor[HTML]{f0f0f0}
\multicolumn{13}{l}{\textit{\textbf{Reinforcement Variants}}} \\
GRPO-4B     & \underline{47.33} & 2.55 & 52.30 & 51.67 & 3.65 & 34.10 & 29.41 & 1.50 & \underline{32.80} & \underline{42.80} & 2.57 & 39.73 \\
DiagAgent-8B~\cite{qiu2025evolving}  & 37.47 & 4.42 & 231.46 & 41.07 & 4.04 & 231.99 & 32.04 & 3.75 & 318.07 & 36.86 & 4.07 & 260.51 \\
DiagAgent-14B~\cite{qiu2025evolving} & 41.33 & 6.16 & 191.81 & 50.73 & 5.52 & 192.11 & \textbf{34.24} & 5.32 & 263.17 & 42.10 & 5.67 & 215.70 \\
\rowcolor[HTML]{ecf0ff}
CDPR-4B (Ours)  & \textbf{53.07} & 2.10 & \textbf{36.65} & \textbf{59.13} & 3.02 & \textbf{17.33} & 30.06 & \underline{1.24} & \textbf{16.58} & \textbf{47.42} & 2.12 & \textbf{23.52} \\
\bottomrule
\end{tabular}
\end{table*}

\noindent\textbf{Baselines.}
We compare against four groups of baselines that cover the spectrum of methods considered in this work. (1) \textbf{Closed-Source LLMs}: GPT-5.4, Claude-Sonnet-4.6, Gemini-3.1-Flash, and DeepSeek-V4, included as a reference for the upper-end capability of generalist models. (2) \textbf{Open-Source LLMs}: general-domain Qwen3-4B/8B/32B \cite{yang2025qwen3}, and Ministral-3-8B \cite{liu2026ministral}, together with medical-domain Med-Gemma-27B \cite{sellergren2025medgemma}, Huatuo-o1-8B \cite{chen2024huatuogpto1medicalcomplexreasoning}, and Baichuan-M1-32B \cite{wang2025baichuan}. (3) \textbf{Medical Agents}: MedAgent~\cite{tang2024medagents}, MDAgent~\cite{kim2024mdagents}, and ToR~\cite{peng2025tree}, three recent agentic frameworks specialized for clinical diagnosis. (4) \textbf{Reinforcement Variants}: GRPO trained on the same Qwen3-4B backbone, and the dynamic diagnostic agent DiagAgent-8B/14B~\cite{qiu2025evolving}.

\noindent\textbf{Metrics.}
We report following metrics: (1) \textbf{Diagnostic Accuracy (Acc.)} is the percentage of cases whose final diagnosis matches the ground truth. (2) \textbf{Average Number of Examinations (AEN)} is the average number of examinations issued per case and reflects diagnostic efficiency. (3) \textbf{Average Exam Cost (AEC)} is the average total monetary cost of examinations per case and reflects economic efficiency. Higher Acc.\ and lower AEN/AEC are better. (4) \textbf{Expert rating of intermediate decisions}: three board-certified physicians independently score randomly sampled intermediate decision nodes on a $1$--$5$ Likert scale along five dimensions --- \textbf{Information Completeness}, \textbf{Evidence Support}, \textbf{Disease Relevance}, \textbf{Diagnostic Gain}, and \textbf{Cost-effectiveness}; the full rubric and protocol are deferred to the appendix.

\noindent\textbf{Implementation Details.}
We use Qwen3-4B-Instruct as the backbone for all training variants. To process examination actions, we first extract the examination action from the model response using regular-expression rules. The extracted action is then matched to the case-specific examination list through both regex matching and LLM-based semantic matching. The same extracted action token is used to compute action entropy, which guides the selection of high-uncertainty decision states. Any action that cannot be parsed is treated as invalid and incurs the unavailable-action penalty. Diagnosis correctness is judged by an LLM-as-a-Judge using GPT-5.4. Full hyperparameters, generation settings, and training configuration are deferred to the appendix.

\subsection{Main Results}

Table~\ref{tab:main_results} summarizes the diagnostic accuracy and examination efficiency of our method against other baselines, on the MIMIC-IV test set and two out-of-domain benchmarks (ClinicalBench, Private). For each method we report Diagnostic Accuracy (Acc., \%), Average Number of Examinations (AEN), and Average Exam Cost (AEC, USD), as well as the cross-dataset average. From this table we draw the following findings:

\noindent\textbf{(1) CDPR-4B achieves the best average accuracy and lowest average exam cost using a 4B backbone.} On the cross-dataset average, CDPR-4B reaches an Acc of $47.42$, surpassing the closed-source GPT-5.4 ($45.30$), the larger general-purpose Qwen3-32B ($42.24$), the medical specialist Med-Gemma-27B ($40.88$), and the dedicated diagnostic agent DiagAgent-14B ($42.10$). At the same time, its average AEC of $23.52$ is the lowest in the entire table, demonstrating that process-level supervision allows a 4B policy to deliver both higher accuracy and substantially lower examination cost than baselines that are several times larger. Looking at individual datasets, the accuracy of CDPR-4B on the Private set ($30.06$) is slightly below DiagAgent-14B ($34.24$); we attribute this to the fact that Private is the hardest benchmark and contains many complex cases whose definitive diagnosis depends on aggregating heterogeneous examinations, where DiagAgent tend to issue more tests pays off in coverage. However, this advantage is bought at a steep price: DiagAgent-14B's AEC on Private is $263.17$ USD against CDPR's $16.58$ USD, so CDPR trades about four points of Acc for roughly $1/16$ of the examination cost, which is the more relevant trade-off in real clinical deployment.

\noindent\textbf{(2) End-to-end RL outperforms multi-agent deliberation on sequential diagnosis.} The three medical-agent baselines, MedAgent, MDAgent, and ToR, attain average Acc of $37.93$, $38.27$, and $39.75$ respectively, with average AEC ranging from $50.97$ to $61.97$ USD; their accuracy sits in the same band as the general-purpose Qwen3 family rather than meaningfully exceeding it. This indicates that on sequential diagnosis, the gain from multiple LLMs deliberating over each step is smaller than the gain from explicitly training a single policy to make decisions, and the deliberation itself amplifies examination cost because every round revisits the trajectory and tends to request additional confirmatory tests.

\noindent\textbf{(3) Medical-domain pre-training does not transfer to active examination ordering.} Huatuo-o1-8B, despite being pre-trained on medical corpora, only reaches an average Acc of $19.07$, far below the general-purpose Qwen3-8B ($37.74$). Med-Gemma-27B ($40.88$) merely matches the general-purpose Qwen3-32B ($42.24$) at the same scale tier. This shows that the strengths conferred by static medical question-answering corpora do not automatically transfer to dynamic diagnosis, where the model must actively decide which examination to issue under cost constraints, and reinforces the need for an explicit process-level reward and reinforcement learning loop, which is precisely what CDPR provides.

\subsection{Ablation Studies}

\begin{table}[!t]
\centering
\caption{Ablation studies on the components of CDPR with Qwen3-4B as the backbone on the in-domain MIMIC-IV test set. The best results are in \textbf{bold}.}
\label{tab:ablation}
\setlength\tabcolsep{6pt}
\begin{tabular}{lccc}
\toprule
\textbf{Method} & \textbf{Acc.\ (\%)} & \textbf{AEN} & \textbf{AEC (USD)} \\
\midrule
\rowcolor[HTML]{ecf0ff}
CDPR (Ours)                     & \textbf{53.07} & \textbf{2.10} & \textbf{36.65} \\
\midrule
~~w/o $R_{\text{proc}}$         & 47.33 & 2.55 & 52.30 \\
~~w/o $\lambda_{\text{test}}$   & 51.85 & 4.10 & 78.40 \\
~~w/o $\lambda_{\text{cost}}$   & 52.10 & 2.20 & 79.20 \\
~~w/o $\lambda_{\text{na}}$     & 50.62 & 4.85 & 62.50 \\
\bottomrule
\end{tabular}
\end{table}

Table~\ref{tab:ablation} reports the contribution of each component in CDPR. We highlight the following observations. (1) Removing the process reward $R_{\text{proc}}$ collapses CDPR back to the outcome-only GRPO baseline, eliminating the dense step-level credit-assignment signal at high-uncertainty decision points; this yields the largest accuracy drop ($53.07\to 47.33$) and a noticeable rise in both AEN and AEC, confirming that $R_{\text{proc}}$ is the core driver of CDPR's gains. (2) Removing $\lambda_{\text{test}}$, $\lambda_{\text{cost}}$, or $\lambda_{\text{na}}$ leaves accuracy comparatively close to the full model but distorts the cost--effort profile in interpretable ways: $\lambda_{\text{test}}$ mainly bounds the number of issued examinations (AEN nearly doubles from $2.10$ to $4.10$), $\lambda_{\text{cost}}$ mainly bounds monetary spend (AEC roughly doubles to $79.20$ even though AEN is unchanged), and $\lambda_{\text{na}}$ guards against clinically infeasible requests, forcing additional retries.

\subsection{Training Efficiency Analysis}

\begin{figure}[!t]
\centering
\includegraphics[width=\linewidth]{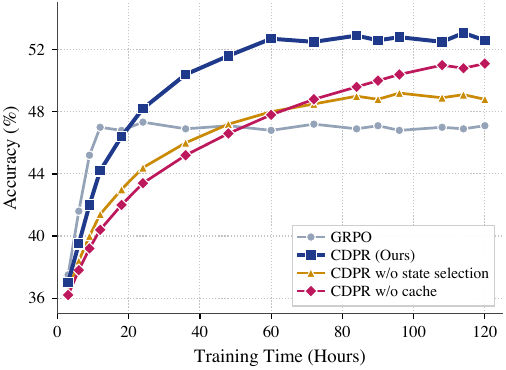}
\caption{Training efficiency comparison on Qwen3-4B. We track diagnostic accuracy on the MIMIC-IV validation split as a function of wall-clock training time (Hours).}
\label{fig:training_efficiency}
\end{figure}

Figure~\ref{fig:training_efficiency} compares four training schemes under the same Qwen3-4B backbone and identical hardware budget: GRPO, CDPR w/o state selection (process reward computed at every state), CDPR w/o cache (rollouts not reused across iterations), and the full CDPR. We make three observations.
(1) GRPO converges earliest but plateaus at the lowest accuracy ($\sim$47\%), confirming that without process supervision the policy cannot resolve credit at decision-critical states.
(2) CDPR w/o cache reaches a competitive plateau but is the slowest, since every selected state launches fresh counterfactual rollouts; this isolates the cache as an efficiency component rather than a performance one.
(3) Full CDPR attains the highest accuracy ($\sim$53.07\%); thanks to state selection and rollout cache, its convergence speed is substantially closer to GRPO than the two ablated variants.

\subsection{Hyperparameter Sensitivity}

\begin{figure}[!t]
\centering
\includegraphics[width=\linewidth]{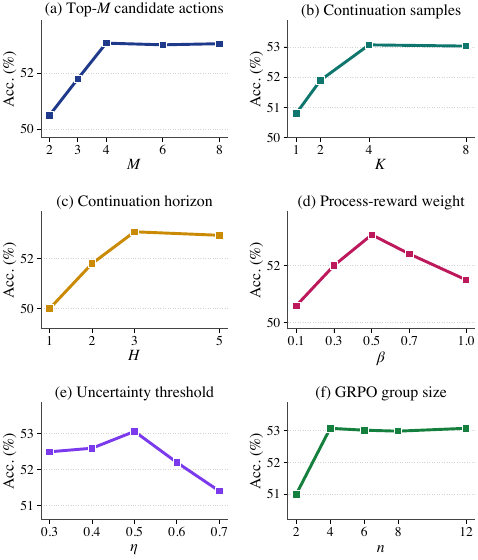}
\caption{Hyperparameter sensitivity on Qwen3-4B. Each panel sweeps one hyperparameter while fixing the others at their defaults, and reports Diagnostic Accuracy (\%).}
\label{fig:sensitivity}
\end{figure}

Figure~\ref{fig:sensitivity} sweeps the six hyperparameters with all others held at their defaults. The hyperparameters control distinct aspects of CDPR: $M$ is the number of top-ranked candidate actions retained at each high-uncertainty state; $K$ is the number of continuation samples drawn per candidate action; $H$ is the continuation horizon, i.e., the maximum number of decision steps the model rolls out per branch; $\beta$ is the global mixing weight of the process reward into the terminal reward; $\eta$ is the uncertainty threshold that selects which states receive process-level supervision; and $n$ is the GRPO group size used for advantage normalization. As shown in Figure~\ref{fig:sensitivity}, accuracy peaks at $M{=}4$, $K{=}4$, $H{=}3$, $\beta{=}0.5$, $\eta{=}0.5$, and $n{=}4$, and we adopt these values as defaults throughout the paper.

\subsection{Expert Evaluation of Intermediate Decisions}

End-task accuracy alone cannot tell whether a method reaches the right diagnosis through a clinically sound trajectory. To probe intermediate decision quality, we ran a blinded expert study: three board-certified physicians independently rated $300$ randomly sampled intermediate decision states ($100$ from each test set) on five dimensions, including Information Completeness, Evidence Support, Disease Relevance, Diagnostic Gain, and Cost-effectiveness. For each sampled state all five compared methods (CDPR, GPT-5.4, Qwen3-32B, Med-Gemma-27B, MDAgent) are evaluated on the same state, so the comparisons within a state are fully paired. Per-state per-method scores are averaged across raters; the full rubric and protocol, including the paired Wilcoxon signed-rank test against each baseline, are given in the appendix.

\begin{figure}[!t]
\centering
\includegraphics[width=\linewidth]{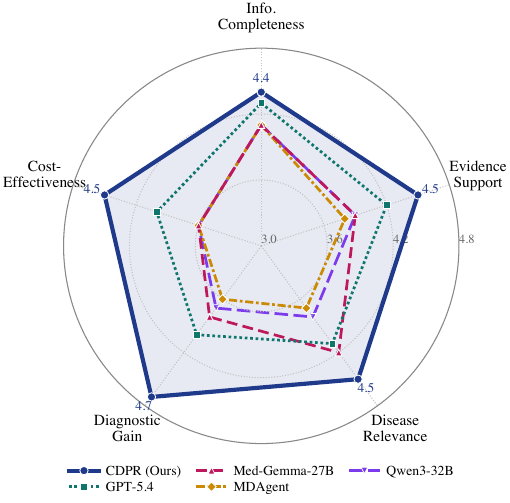}
\caption{Expert evaluation of intermediate decisions across five dimensions. Scores are 1--5 Likert ratings averaged over three physicians and 300 sampled decision states; for each state, all five methods are evaluated on the same state with model identity blinded. Markers indicate dimensions on which CDPR is significantly above the next-best baseline (paired Wilcoxon signed-rank test, $p<0.05$ after Holm correction).}
\label{fig:human_eval}
\end{figure}

Figure~\ref{fig:human_eval} compares CDPR against four strong baselines. CDPR leads on all five dimensions, with the largest margin on Diagnostic Gain ($4.7$ vs.\ next-best $4.0$, paired Wilcoxon Holm-adjusted $p<0.001$), directly validating the uncertainty-driven state selection: by densifying supervision at high-entropy decision points, CDPR learns to pick actions that most reduce diagnostic uncertainty. CDPR also tops Cost-effectiveness ($4.5$ vs.\ $4.0$, Holm-adjusted $p<0.001$), reflecting the explicit cost penalties in the utility function. Med-Gemma-27B outperforms GPT-5.4 on Disease Relevance ($4.2$ vs.\ $4.1$) thanks to its medical pretraining, but trails on Information Completeness and Evidence Support, where general-purpose reasoning still helps. Overall, the radar profile confirms that CDPR's gains in end-task accuracy come from genuinely better stepwise decisions.

\subsection{Dynamic vs.\ Static Diagnosis}

\begin{table}[!t]

\setlength\tabcolsep{6pt}
\centering
\caption{CDPR (dynamic diagnosis) versus static diagnosis on the in-domain MIMIC-IV test set, both built on the Qwen3-4B backbone. Static diagnosis is fed the entire case in a single shot, while CDPR sequentially issues examinations under cost constraints. Numbers in parentheses denote the relative reduction over static diagnosis.}
\begin{tabular}{lcc}
\toprule
\textbf{Method} & \textbf{Acc.\ (\%)} & \textbf{AEC (USD)} \\
\midrule
Static diagnosis           & 50.23 & 395.40 \\
\rowcolor[HTML]{ecf0ff}
CDPR (Ours)                & \textbf{53.07} & \textbf{36.65}~\textcolor[HTML]{15803d}{$\blacktriangledown$\,($-$90.7\%)} \\
\bottomrule
\label{tab:dynamic_vs_static}
\end{tabular}
\end{table}

Table~\ref{tab:dynamic_vs_static} compares CDPR with a static-diagnosis baseline that receives the full case description in one shot. This setting is unfair to CDPR, since CDPR must actively decide which examinations to issue rather than seeing all evidence at once. Despite this disadvantage, CDPR matches and slightly surpasses static diagnosis in accuracy ($53.07$ vs.\ $50.23$) while spending only about one tenth of its examination cost ($36.65$ vs.\ $395.40$ USD, a $90.7\%$ reduction). This confirms that CDPR retains diagnostic competence under sequential, cost-aware constraints and achieves a markedly better accuracy-cost trade-off, validating its efficiency advantage in realistic dynamic-diagnosis situations.


\section{Conclusion}
In this paper, we propose CDPR, a counterfactual action-value process reward for cost-aware sequential medical diagnosis. CDPR targets high-uncertainty decision states, evaluates the chosen action against counterfactual alternatives by short rollouts, and transforms outcome differences into step-level rewards. By reusing training trajectories with a rollout cache, CDPR further reduces the cost of counterfactual evaluation. Experiments on one in-domain and two out-of-domain medical diagnosis benchmarks show that CDPR delivers higher diagnostic accuracy and substantially lower examination cost than other methods. By turning sparse outcome rewards into dense step-level supervision under a tight rollout budget, CDPR addresses the core credit-assignment difficulty in cost-aware sequential medical diagnosis.

\bibliography{aaai2026}

\clearpage

\clearpage
\appendix

\begin{onecolumn}
\section{Appendix}

\subsection{Dataset Description}

\noindent\textbf{MIMIC-IV.}
MIMIC-IV is a large-scale, publicly accessible critical-care database that contains de-identified electronic health records collected at a tertiary academic medical center, primarily covering intensive-care and emergency-department admissions. We use it as our in-domain corpus and follow the data-construction pipeline of DiagGym~\cite{qiu2025evolving}: each discharge summary is parsed into a structured patient state, namely the chief complaint, history of present illness, past medical history, and physical examination, together with the chronologically ordered laboratory and imaging examinations, their results, and the final discharge diagnosis. The resulting cases predominantly cover critical-care and emergency-medicine conditions, including cardiovascular events, sepsis and other infectious processes, gastrointestinal and hepatobiliary disorders, acute renal injury, respiratory failure, and neurological emergencies. We use $23{,}377$ cases for training and hold out $750$ further cases as the in-domain test set.

\noindent\textbf{ClinicalBench.}
ClinicalBench~\cite{yan2026clinicallab} is an end-to-end multi-departmental clinical diagnostic benchmark constructed from real-world electronic medical records donated by Chinese tertiary hospitals, with strict de-identification and quality control. It contains $1{,}500$ patient cases spanning $24$ clinical departments, ranging from internal medicine, cardiology, neurology, gastroenterology, and oncology to surgery, obstetrics and gynecology, pediatrics, and otolaryngology, and covers roughly $150$ distinct diseases. Each case preserves the complete diagnostic trajectory: preliminary reception, history taking, ordered examinations and their reports, differential reasoning, and final diagnosis. Because its source distribution differs substantially from MIMIC-IV, we use the entire $1{,}500$-case set as a held-out out-of-domain test set; no part of it is used during training.

\noindent\textbf{Private Dataset.}
The Private Dataset is curated in collaboration with a partner tertiary  hospital and consists of $952$ fully de-identified inpatient cases. Every case spans the complete admission-to-discharge trajectory and is annotated with chief complaint, history of present illness, past medical history, physical examination, laboratory and imaging examinations together with their reports, and the discharge diagnosis. The disease distribution exhibits a long-tailed shape and spans a broad spectrum of common inpatient conditions across multiple organ systems: hepatobiliary disease (e.g., hepatic cyst, hepatic hemangioma), cardiovascular disease (e.g., hypertension at multiple risk grades, aortic sclerosis), endocrine and metabolic disorders (e.g., electrolyte disturbances, nodular goiter), genitourinary and renal conditions (e.g., bilateral/unilateral renal cysts, renal calculi), gastrointestinal disease (e.g., chronic gastritis with or without erosion, intestinal obstruction), and respiratory disease (e.g., obstructive pneumonia, pulmonary nodules). The corpus thus complements MIMIC-IV with hospital admissions whose acuity profile and disease mix differ from those of intensive- and emergency-care units. We use the entire dataset as a second held-out out-of-domain test set.

\noindent\textbf{Unified examination cost.}
To enable cost-aware learning and evaluation, we attach a unified U.S.\ dollar (USD) cost to every examination item across the three datasets, with values drawn from publicly available U.S.\ medical fee schedules. Laboratory tests are priced using the CMS Clinical Laboratory Fee Schedule (CLFS)\footnote{\url{https://www.cms.gov/medicare/payment/fee-schedules/clinical-laboratory-fee-schedule-clfs}}; imaging, electrocardiography, electroencephalography, echocardiography, and other non-laboratory diagnostic procedures are priced using the CMS Physician Fee Schedule (PFS)\footnote{\url{https://www.cms.gov/medicare/payment/fee-schedules/physician}}; facility-heavy outpatient procedures are priced using the CMS Outpatient Prospective Payment System Addendum B\footnote{\url{https://www.cms.gov/medicare/payment/prospective-payment-systems/hospital-outpatient-pps/quarterly-addenda-updates}}. To avoid double-charging when multiple sub-items belong to the same billing event, we further introduce a panel-aware billing-group map that consolidates fine-grained keys into parent groups: for example, \texttt{BLOOD Na}, \texttt{BLOOD K}, \texttt{BLOOD Cl}, \texttt{BLOOD Creat}, \texttt{BLOOD Glucose}, and \texttt{BLOOD UreaN} share one \texttt{BMP\_PANEL}; \texttt{WBC}, \texttt{Hgb}, \texttt{Hct}, and \texttt{Platelet} share one \texttt{CBC\_PANEL}; and urine micro-tests share one \texttt{UA\_PANEL}. At runtime, ordering any sub-item charges its parent billing group exactly once per case. The same cost map and billing-group rules are applied during both training and inference, and the cumulative monetary budget is strictly enforced.

\noindent\textbf{Unified data format.}
We process all three datasets into a single record format so that the agent interacts with them through identical interfaces. Each record contains a case summary (chief complaint, present illness, past medical history, and personal, family, and allergy histories), a key-pertinent-results dictionary (physical examinations together with all laboratory and imaging findings), the final diagnosis, and the per-case examination cost map in USD. Table~\ref{tab:app_case} shows one sample record drawn from MIMIC-IV.

\begin{longtable}{p{\linewidth}}
\caption{A case in the MIMIC-IV dataset.}
\label{tab:app_case}\\
\toprule
\endfirsthead
\multicolumn{1}{c}{{\tablename\ \thetable{} -- Continued from previous page}} \\
\toprule
\endhead
\midrule
\multicolumn{1}{r}{{Continued on next page}} \\
\endfoot
\bottomrule
\endlastfoot

\textbf{``note\_id'':} ``19449006-DS-18'',\\
\textbf{``case\_summary'':} ``- Patient Information: \_\_\_ year old female\textbackslash n- Chief Complaint: Abdominal pain\textbackslash n- History of Present Illness: The patient reports gradual onset right lower quadrant (RLQ) pain with radiation to the back, accompanied by nausea and a small amount of diarrhea. She denies hematuria. The pain is described as sharp, moderate in severity, and sudden in onset. She states it feels similar to prior episodes of renal colic.\textbackslash n- Past Medical History: Knee surgery and one episode of nephrolithiasis\textbackslash n- Personal History: None\textbackslash n- Family History: Non-contributory\textbackslash n- Allergy History: No Known Allergies / Adverse Drug Reactions'',\\
\textbf{``key\_pertinent\_results\_dict'':} \{\\
\quad \textbf{``Vital Signs (Physical Examination)'':} ``Temp: 99.2, HR: 80, BP: 122/81, Resp: 14, O(2)Sat: 100'',\\
\quad \textbf{``Constitutional (Physical Examination)'':} ``Comfortable'',\\
\quad \textbf{``HEENT (Physical Examination)'':} ``Normocephalic, atraumatic, Pupils equal, round and reactive to light, Extraocular muscles intact, Oropharynx within normal limits'',\\
\quad \textbf{``Chest (Physical Examination)'':} ``Clear to auscultation'',\\
\quad \textbf{``Cardiovascular (Physical Examination)'':} ``Regular Rate and Rhythm, Normal first and second heart sounds'',\\
\quad \textbf{``Abdominal (Physical Examination)'':} ``Soft, Nondistended, TTP mcburney's point'',\\
\quad \textbf{``GU/Flank (Physical Examination)'':} ``No costovertebral angle tenderness'',\\
\quad \textbf{``Extr/Back (Physical Examination)'':} ``No cyanosis, clubbing or edema'',\\
\quad \textbf{``Skin (Physical Examination)'':} ``No rash, Warm and dry'',\\
\quad \textbf{``Neuro (Physical Examination)'':} ``Speech fluent'',\\
\quad \textbf{``Psych (Physical Examination)'':} ``Normal mood, Normal mentation'',\\
\quad \textbf{``Additional Findings (Physical Examination)'':} ``No petechiae'',\\
\quad \textbf{``URINE CULTURE'':} ``Organism: ESCHERICHIA COLI; Isolate Number: 1.0; Antibiotic: AMPICILLIN (Dilution Text: 8, Comparison: =, Value: 8.0); Interpretation: S; Comments: Culture workup discontinued. Further incubation showed contamination with mixed skin/genital flora. Clinical significance of isolate(s) uncertain. Interpret with caution.'',\\
\quad \textbf{``Bacteria'':} ``Comments: FEW.'',\\
\quad \textbf{``Urine Analysis'':} ``Bilirubin: Units: mg/dL; Comments: NEG.; Epithelial Cells: Value: 17; Numeric Value: 17.0; Units: \#/hpf; Glucose: Units: mg/dL; Comments: NEG.; Ketone: Units: mg/dL; Comments: TR.; Leukocytes: Comments: SM.; Nitrite: Comments: NEG.; pH: Value: 6.5; Numeric Value: 6.5; Units: units; Reference Range: 5.0--8.0; Protein: Value: 30; Numeric Value: 30.0; Units: mg/dL; RBC: Value: 1; Numeric Value: 1.0; Units: \#/hpf; Reference Range: 0.0--2.0; Specific Gravity: Value: 1.024; Numeric Value: 1.024; Reference Range: 1.001--1.035; Urine Appearance: Comments: Hazy.; Urine Color: Comments: Yellow.; Urine Mucous: Comments: MANY.; Urobilinogen: Value: 2; Numeric Value: 2.0; Units: mg/dL; Reference Range: 0.2--1.0; Flag: abnormal; WBC: Value: 5; Numeric Value: 5.0; Units: \#/hpf; Reference Range: 0.0--5.0'',\\
\quad \textbf{``Blood'':} ``Comments: NEG.'',\\
\quad \textbf{``Yeast'':} ``Value: NONE'',\\
\quad \textbf{``Urine Pregnancy Test'':} ``HCG, Urine, Qualitative: Units: +/-; Comments: NEGATIVE. FOR QUANTITATION OF POSITIVES, SEND SERUM FOR HCG.'',\\
\quad \textbf{``Length of Urine Collection'':} ``Comments: RANDOM.'',\\
\quad \textbf{``Anion Gap'':} ``Value: 18; Numeric Value: 18.0; Units: mEq/L; Reference Range: 8.0--20.0'',\\
\quad \textbf{``Comprehensive Metabolic Panel'':} ``Bicarbonate: Value: 24; Numeric Value: 24.0; Units: mEq/L; Reference Range: 22.0--32.0'',\\
\quad \textbf{``Kidney Function Tests'':} ``Chloride: Value: 101; Numeric Value: 101.0; Units: mEq/L; Reference Range: 96.0--108.0; Creatinine: Value: 0.9; Numeric Value: 0.9; Units: mg/dL; Reference Range: 0.4--1.1; Glucose: Numeric Value: 149.0; Units: mg/dL; Reference Range: 70.0--100.0; Flag: abnormal; Comments: IF FASTING, 70--100 NORMAL, $>$125 PROVISIONAL DIABETES.; Potassium: Numeric Value: 4.8; Units: mEq/L; Reference Range: 3.3--5.1; Comments: HEMOLYSIS FALSELY ELEVATES K.; Sodium: Value: 138; Numeric Value: 138.0; Units: mEq/L; Reference Range: 133.0--145.0; Urea Nitrogen: Value: 17; Numeric Value: 17.0; Units: mg/dL; Reference Range: 6.0--20.0'',\\
\quad \textbf{``Estimated GFR'':} ``Estimated GFR (MDRD equation): Comments: Using this patient's age, gender, and serum creatinine value of 0.9. Estimated GFR = 67 if non African-American (mL/min/1.73 m$^2$). Estimated GFR $>$ 75 if African-American (mL/min/1.73 m$^2$). For comparison, mean GFR for age group 40--49 is 99 (mL/min/1.73 m$^2$). GFR $<$ 60 = Chronic Kidney Disease, GFR $<$ 15 = Kidney Failure.'',\\
\quad \textbf{``Light Green Top Hold'':} ``Comments: HOLD.'',\\
\quad \textbf{``Liver Function Test'':} ``INR(PT): Value: 0.9; Numeric Value: 0.9; Reference Range: 0.9--1.1; PT: Numeric Value: 9.7; Units: sec; Reference Range: 9.4--12.5; Comments: VERIFIED.'',\\
\quad \textbf{``Coagulation Profile'':} ``PTT: Numeric Value: 23.2; Units: sec; Reference Range: 25.0--36.5; Flag: abnormal; Comments: VERIFIED.'',\\
\quad \textbf{``Complete Blood Count'':} ``Basophils: Value: 0.4; Numeric Value: 0.4; Units: \%; Reference Range: 0.0--2.0; Eosinophils: Value: 0.6; Numeric Value: 0.6; Units: \%; Reference Range: 0.0--4.0; Hematocrit: Value: 46.7; Numeric Value: 46.7; Units: \%; Reference Range: 36.0--48.0; Hemoglobin: Value: 16.1; Numeric Value: 16.1; Units: g/dL; Reference Range: 12.0--16.0; Flag: abnormal; Lymphocytes: Value: 6.5; Numeric Value: 6.5; Units: \%; Reference Range: 18.0--42.0; Flag: abnormal; MCH: Value: 32.3; Numeric Value: 32.3; Units: pg; Reference Range: 27.0--32.0; Flag: abnormal; MCHC: Value: 34.6; Numeric Value: 34.6; Units: \%; Reference Range: 31.0--35.0; Monocytes: Value: 4.5; Numeric Value: 4.5; Units: \%; Reference Range: 2.0--11.0; Neutrophils: Value: 88.0; Numeric Value: 88.0; Units: \%; Reference Range: 50.0--70.0; Flag: abnormal; Platelet Count: Value: 273; Numeric Value: 273.0; Units: K/uL; Reference Range: 150.0--440.0; RDW: Value: 12.9; Numeric Value: 12.9; Units: \%; Reference Range: 10.5--15.5; Red Blood Cells: Value: 5.00; Numeric Value: 5.0; Units: m/uL; Reference Range: 4.2--5.4; White Blood Cells: Value: 18.5; Numeric Value: 18.5; Units: K/uL; Reference Range: 4.0--11.0; Flag: abnormal'',\\
\quad \textbf{``Mean Corpuscular Volume'':} ``MCV: Value: 93; Numeric Value: 93.0; Units: fL; Reference Range: 82.0--98.0'',\\
\quad \textbf{``Green Top Hold, plasma'':} ``Green Top Hold (plasma): Value: HOLD. DISCARD GREATER THAN 4 HOURS OLD.'',\\
\quad \textbf{``Lactate'':} ``Value: 2.2; Numeric Value: 2.2; Units: mmol/L; Reference Range: 0.5--2.0; Flag: abnormal'',\\
\quad \textbf{``CT of abdomen and pelvis'':} ``Preliminary Report 1. Acute appendicitis.''\\
\},\\
\textbf{``final\_diagnosis'':} ``Acute appendicitis. The diagnosis is supported by the patient's clinical presentation of RLQ pain, nausea, and diarrhea, along with laboratory findings of leukocytosis and neutrophilia, elevated lactate, and CT imaging confirming a dilated, fluid-filled appendix.'',\\
\textbf{``diagnosis\_results'':} ``Acute appendicitis'',\\
\textbf{``exam\_cost\_map (USD)'':} \{\\
\quad \textbf{``Vital Signs (Physical Examination)'':} 0.00,\\
\quad \textbf{``Constitutional (Physical Examination)'':} 0.00,\\
\quad \textbf{``HEENT (Physical Examination)'':} 0.00,\\
\quad \textbf{``Chest (Physical Examination)'':} 0.00,\\
\quad \textbf{``Cardiovascular (Physical Examination)'':} 0.00,\\
\quad \textbf{``Abdominal (Physical Examination)'':} 0.00,\\
\quad \textbf{``GU/Flank (Physical Examination)'':} 0.00,\\
\quad \textbf{``Extr/Back (Physical Examination)'':} 0.00,\\
\quad \textbf{``Skin (Physical Examination)'':} 0.00,\\
\quad \textbf{``Neuro (Physical Examination)'':} 0.00,\\
\quad \textbf{``Psych (Physical Examination)'':} 0.00,\\
\quad \textbf{``Additional Findings (Physical Examination)'':} 0.00,\\
\quad \textbf{``URINE CULTURE'':} 8.50,\\
\quad \textbf{``Bacteria'':} 5.00,\\
\quad \textbf{``Urine Analysis'':} 5.00,\\
\quad \textbf{``Blood'':} 5.00,\\
\quad \textbf{``Yeast'':} 5.00,\\
\quad \textbf{``Urine Pregnancy Test'':} 5.00,\\
\quad \textbf{``Length of Urine Collection'':} 5.00,\\
\quad \textbf{``Anion Gap'':} 8.46,\\
\quad \textbf{``Comprehensive Metabolic Panel'':} 10.56,\\
\quad \textbf{``Kidney Function Tests'':} 8.68,\\
\quad \textbf{``Estimated GFR'':} 8.68,\\
\quad \textbf{``Light Green Top Hold'':} 5.00,\\
\quad \textbf{``Liver Function Test'':} 5.00,\\
\quad \textbf{``Coagulation Profile'':} 18.00,\\
\quad \textbf{``Complete Blood Count'':} 6.47,\\
\quad \textbf{``Mean Corpuscular Volume'':} 5.00,\\
\quad \textbf{``Green Top Hold, plasma'':} 5.00,\\
\quad \textbf{``Lactate'':} 11.57,\\
\quad \textbf{``CT of abdomen and pelvis'':} 29.73\\
\},\\
\end{longtable}

\subsection{Prompt Templates}
\label{app:prompts}

This section shows the prompts used in our framework, each driving a distinct module of the sequential-diagnosis pipeline. The \emph{Sequential Diagnosis Agent Prompt} (Figure~\ref{fig:prompt_agent}) is the policy prompt: it instructs the actor to maintain an explicit short list of differential diagnoses at every step, choose a single examination that best discriminates among them, and commit to a final diagnosis once the evidence is sufficient. The \emph{Exam Key Matching Prompt} (Figure~\ref{fig:prompt_exam}) is a lightweight terminology-alignment module: it maps a free-form examination request from the agent to one of the available examination keys in the case record, so that the environment can return the matching result without leaking values. The \emph{Diagnosis Evaluation Prompt} (Figure~\ref{fig:prompt_judge}) is the outcome judge: given a model prediction and the ground-truth diagnosis, it returns a single binary verdict ``Correct''/``Wrong'' that defines the terminal accuracy reward.

\begin{center}
\begin{tcolorbox}[breakable, title=\textbf{Sequential Diagnosis Agent Prompt},
                  colback=gray!5!white, colframe=gray!50!black, fonttitle=\bfseries,
                  before skip=8pt, after skip=8pt, fontupper=\small]
You are a medical AI assistant performing sequential diagnosis. You are given a patient's initial clinical information and must iteratively request medical examinations/tests to gather more information, then provide a final diagnosis when you have enough evidence.

\textbf{Core Strategy}
\begin{itemize}[leftmargin=*, nosep]
    \item Start broad: use foundational labs to narrow the differential before committing to specialized tests.
    \item Be efficient: each test should meaningfully change your differential or confirm/exclude a leading diagnosis.
    \item Be specific: when diagnosing, name the precise condition (e.g., ``Community-acquired pneumonia'' not just ``Lung infection'').
\end{itemize}

\textbf{Rules}
\begin{enumerate}[leftmargin=*, nosep]
    \item At each step, explicitly list your top 2--3 differential diagnoses before deciding the next action.
    \item Choose the test that best discriminates between your leading differentials.
    \item Request only ONE specific test per turn.
    \item Once key discriminating evidence is obtained, commit to a final diagnosis --- do not over-investigate.
    \item Use standard medical terminology for test names.
\end{enumerate}

\textbf{Available Examinations} (representative examples) \\
\textit{Laboratory:} Complete Blood Count, Comprehensive Metabolic Panel, Liver Function Test, Kidney Function Tests, Urine Analysis, Coagulation Profile, Troponin, Lactate, Blood Culture, Urine Culture, Arterial Blood Gas, Lipase, D-dimer, Thyroid Function Tests, Blood Glucose, HbA1c, Procalcitonin, BNP/NT-proBNP, Iron Studies, Vitamin B12/Folate. \\
\textit{Imaging \& Procedures:} Chest X-ray, CT Abdomen and Pelvis, CT Head, CT Chest, Abdominal Ultrasound, Echocardiography, ECG, MRI Brain, Lumbar Puncture, Endoscopy.

\textbf{Common Diagnoses in This Setting} \\
Pneumonia, Urinary tract infection, Acute appendicitis, Cellulitis, Acute pancreatitis, Acute cholecystitis, Heart failure, Subarachnoid hemorrhage, Acute kidney injury, Upper GI bleed, Sepsis, Pulmonary embolism, Diabetic ketoacidosis, Acute myocardial infarction, Stroke, Bowel obstruction, Deep vein thrombosis.

\textbf{Response Format} \\
If you need more information (request a test): \\
\texttt{THINKING: [Top 2--3 differential diagnoses ranked by likelihood, and why this test best discriminates between them]} \\
\texttt{ACTION: REQUEST\_TEST} \\
\texttt{Test needed: [one specific test/examination name]} \\
\texttt{Reason: [why this test is needed]}

If you have enough information (final diagnosis): \\
\texttt{THINKING: [Brief summary of key findings supporting the diagnosis]} \\
\texttt{ACTION: FINAL\_DIAGNOSIS} \\
\texttt{Diagnosis: [your final diagnosis]} \\
\texttt{Reason: [clinical reasoning supporting the diagnosis]}

\textbf{Example} (Requesting imaging) \\
\texttt{THINKING:} Patient presents with fever, productive cough, and crackles on lung auscultation. Top differentials: (1) Community-acquired pneumonia, (2) Bronchitis, (3) Atypical pneumonia. Chest X-ray will best discriminate by detecting pulmonary infiltrates. \\
\texttt{ACTION: REQUEST\_TEST} \\
\texttt{Test needed:} Chest X-ray \\
\texttt{Reason:} To confirm pulmonary infiltrates and assess extent of infection.
\end{tcolorbox}
\captionof{figure}{The system prompt that drives the diagnostic agent at every interaction step.}
\label{fig:prompt_agent}
\end{center}

\begin{center}
\begin{tcolorbox}[breakable, title=\textbf{Exam Key Matching Prompt},
                  colback=gray!5!white, colframe=gray!50!black, fonttitle=\bfseries,
                  before skip=8pt, after skip=8pt, fontupper=\small]
You are a medical terminology matching assistant. Given a list of available examination names and a requested test, determine which examination name (if any) matches the request.

\textbf{Rules}
\begin{itemize}[leftmargin=*, nosep]
    \item Match using medical knowledge: abbreviations, synonyms, and alternative names all count (e.g., ``CBC'' = ``Complete Blood Count'', ``ECG'' = ``Electrocardiogram'', ``CXR'' = ``Chest X-ray'').
    \item Return EXACTLY the matching name from the available list --- do not modify it.
    \item If no match exists, output ``\texttt{NO\_MATCH}''.
\end{itemize}

\textbf{Response Format} \\
If matched: output ONLY the exact examination name from the list, nothing else. \\
If not matched: output ONLY ``\texttt{NO\_MATCH}'', nothing else.

\textbf{Examples} \\
Available: \texttt{["Complete Blood Count", "Chest X-ray", "Basic Metabolic Panel"]} \\
Requested: \texttt{CBC} \\
Response: \texttt{Complete Blood Count}

Available: \texttt{["Electrocardiogram", "Troponin I", "Chest X-ray"]} \\
Requested: \texttt{ECG} \\
Response: \texttt{Electrocardiogram}

Available: \texttt{["Complete Blood Count", "Basic Metabolic Panel"]} \\
Requested: \texttt{CT Abdomen} \\
Response: \texttt{NO\_MATCH}

\textbf{User Template} \\
\texttt{Available: \{exam\_keys\}} \\
\texttt{Requested: \{requested\_test\}} \\
\texttt{Response:}
\end{tcolorbox}
\captionof{figure}{The lightweight prompt that aligns the agent's free-form test request with one of the case's available examination keys.}
\label{fig:prompt_exam}
\end{center}

\begin{center}
\begin{tcolorbox}[breakable, title=\textbf{Diagnosis Evaluation (Judge) Prompt},
                  colback=gray!5!white, colframe=gray!50!black, fonttitle=\bfseries,
                  before skip=8pt, after skip=8pt, fontupper=\small]
You are a medical diagnosis evaluation system. Your task is to judge whether a model-predicted diagnosis is clinically correct when compared to the ground truth diagnosis.

\textbf{Evaluation Rules}
\begin{enumerate}[leftmargin=*, nosep]
    \item The same disease may have multiple names or aliases (e.g., ``Heart attack'' = ``Myocardial infarction'' = ``MI'').
    \item Abbreviations and full names are equivalent (e.g., ``COPD'' = ``Chronic Obstructive Pulmonary Disease'').
    \item If the predicted diagnosis captures the core disease correctly but mentions additional complications or details, it is still CORRECT.
    \item If the predicted diagnosis is a broader category that includes the ground truth (e.g., \texttt{pred=}``Cardiac disease'' vs.\ \texttt{gt=}``Acute myocardial infarction''), it is WRONG --- the prediction must be specific enough.
    \item If the predicted diagnosis is a different disease entirely, it is WRONG.
    \item Only evaluate the diagnosis itself --- ignore differences in reasoning, treatment, or symptom descriptions.
\end{enumerate}

\textbf{Response Format} \\
Output ONLY one word: ``\texttt{Correct}'' or ``\texttt{Wrong}''. Do not output any other content.

\textbf{Examples} \\
\textit{Example 1 --- Correct (same disease, different wording):} \\
\texttt{[pred\_diag]:} Acute myocardial infarction \\
\texttt{[gt\_diag]:} Heart attack \\
\texttt{Output:} \texttt{Correct}

\textit{Example 2 --- Correct (prediction includes ground truth with extra detail):} \\
\texttt{[pred\_diag]:} Community-acquired pneumonia with possible sepsis \\
\texttt{[gt\_diag]:} Pneumonia \\
\texttt{Output:} \texttt{Correct}

\textit{Example 3 --- Wrong (different disease):} \\
\texttt{[pred\_diag]:} Gastroesophageal reflux disease \\
\texttt{[gt\_diag]:} Peptic ulcer disease \\
\texttt{Output:} \texttt{Wrong}

\textbf{User Template} \\
\texttt{[pred\_diag]} \\
\texttt{\{pred\_diag\}} \\
\texttt{[gt\_diag]} \\
\texttt{\{gt\_diag\}}
\end{tcolorbox}
\captionof{figure}{The judge prompt that turns the predicted final diagnosis into the binary terminal reward used during training and evaluation.}
\label{fig:prompt_judge}
\end{center}

\subsection{Hyperparameters and Training Details}
\label{app:hyperparameters}

We detail in this section the hyperparameters used for CDPR training and evaluation. We base our implementation on the verl framework and report a typical training configuration. All training runs are performed on $2$ NVIDIA H20 GPUs, with the actor and the frozen stale-actor (used for counterfactual continuation) co-located on the same two devices via separate GPU memory budget. The diagnosis judge model is accessed via the GPT-5.4 API.

\noindent\textbf{Backbone and Rollout.}
The actor is initialized from Qwen3-4B-Instruct-2507 and trained with multi-turn rollouts capped at $8$ assistant turns per trajectory, a maximum prompt length of $2048$ tokens, and a maximum response length of $4096$ tokens. Rollouts are served by SGLang with FP8 quantization, asynchronous mode, and a tensor-parallel size of $2$. The CDPR \emph{stale-actor} that produces counterfactual continuations is hosted by a separate vLLM server initialized from the same Qwen3-4B-Instruct-2507 weights and frozen for the entire run, so continuations remain well-formed regardless of how the main actor drifts during training. Diagnosis correctness is judged by a closed-source GPT-5.4 model accessed via API, using the prompt detailed in Figure~\ref{fig:prompt_judge}; the same judge is used during both training and evaluation to ensure that the scoring criterion remains consistent throughout.

\noindent\textbf{Optimization.}
We optimize the actor with GRPO~\cite{qiu2025evolving}: $n=4$ rollouts per prompt, a global training batch of $16$ prompts (yielding $16\times 4 = 64$ trajectories per iteration), a PPO mini-batch of $16$ and a per-GPU PPO micro-batch of $8$, and a single PPO inner epoch. The actor uses learning rate $5\times 10^{-6}$, bfloat16 parameters with FSDP and gradient checkpointing, and FlashAttention-2; KL regularization to the reference model is disabled (\texttt{use\_kl\_loss=False}, \texttt{use\_kl\_in\_reward=False}). We run $3$ epochs of training and save/evaluate every $200$ optimizer steps. The data loader uses a fixed seed for reproducible ordering.

\noindent\textbf{CDPR-specific Hyperparameters.}
The counterfactual rollout settings follow the notation in the main paper: top-$M$ candidate actions $M=4$, $K=4$ continuations per branch, continuation horizon $H=3$, and $n_s=8$ action samples for uncertainty estimation, with the uncertainty threshold $\eta=0.5$. The terminal-reward shaping coefficients are $\lambda_{\text{test}}=0.05$, $\lambda_{\text{cost}}=0.02$, $\lambda_{\text{na}}=0.10$, and the discount $\gamma=1.0$. The process reward is mixed into the terminal reward with weight $\beta=0.5$ and clipped at $\pm c$ with $c=1.0$. The continuation cache has capacity $8192$ entries.

\noindent\textbf{Evaluation Setup.}
At evaluation time we use SGLang with the same generation length limits as training (max prompt $2048$, max response $4096$). The judge prompt (Figure~\ref{fig:prompt_judge}) returns a binary correctness label, which is averaged over the test set to obtain Accuracy; Average Number of Examinations (AEN) are obtained from the real trajectory and Average Exam Cost (AEC, USD) are computed using the unified USD cost map, respectively. All test sets (in-domain MIMIC-IV, ClinicalBench, Private) are evaluated with the same configuration; no test-time tuning is performed.

\subsection{Expert Evaluation Protocol}
\label{app:expert_eval}

This appendix details the expert evaluation reported in the main text. The study aims to assess the quality of \emph{intermediate} diagnostic decisions, rather than only the final diagnosis.

\noindent\textbf{Participants and Sampling.}
We invited three board-certified physicians with at least five years of clinical experience. From each of the three test sets (MIMIC-IV, ClinicalBench, Private) we randomly sampled $100$ intermediate decision states, yielding $300$ states in total. For each sampled state, all five compared methods (CDPR, GPT-5.4, Qwen3-32B, Med-Gemma-27B, MDAgent) are evaluated on the \emph{same} state: each method's reasoning and proposed next action are collected at that state and presented to the raters in randomized order with model identity blinded. Each rater therefore scores five method outputs per state, conditioned on identical patient information (chief complaint, history, prior findings), so that method comparisons within a state are fully paired. Downstream outcomes and the final diagnosis are withheld to avoid bias.

\noindent\textbf{Scoring Dimensions.}
Each node is rated on a $1$--$5$ Likert scale along five dimensions. The first two assess the \emph{reasoning} quality; the last three assess the proposed \emph{next action}.

\begin{itemize}
    \item \textbf{Information Completeness.} Whether the model fully and accurately exploits the available case state, including symptoms, history, prior results, and key positive/negative findings.
    \begin{itemize}
        \item[$1$:] Clearly misses or misreads key information, breaking the basis of subsequent reasoning.
        \item[$2$:] Uses only part of the information; many important clinical findings are overlooked or misread.
        \item[$3$:] Captures the main information but still misses some important details.
        \item[$4$:] Captures most key information accurately, with only minor omissions.
        \item[$5$:] Comprehensively and accurately captures all key positives and negatives.
    \end{itemize}
    \item \textbf{Evidence Support.} Whether each step of reasoning is grounded in the currently available evidence, with no unwarranted leaps or contradictions.
    \begin{itemize}
        \item[$1$:] Reasoning lacks evidential support or contradicts the current state.
        \item[$2$:] Contains many unsupported inferences, over-interpretations, or weakly grounded claims.
        \item[$3$:] Mostly acceptable but with some weakly supported judgments.
        \item[$4$:] Generally well supported by the current evidence, with only minor gaps.
        \item[$5$:] Every key inference is explicitly traceable to the current evidence.
    \end{itemize}
    \item \textbf{Disease Relevance.} Whether the proposed next action is directed at the current target disease, suspected diseases, or critical differentials.
    \begin{itemize}
        \item[$1$:] Action is essentially unrelated to any plausible disease.
        \item[$2$:] Only indirectly related; weakly aligned with the most likely candidate diseases.
        \item[$3$:] Loosely related to some candidate but not well targeted.
        \item[$4$:] Reasonably aligned with the main suspected diseases or differentials.
        \item[$5$:] Tightly focused on the most likely disease or the most critical differential.
    \end{itemize}
    \item \textbf{Diagnostic Gain.} Whether the action is expected to provide new and valuable diagnostic information --- e.g., reducing uncertainty, discriminating between key differentials, or unblocking the next reasoning step.
    \begin{itemize}
        \item[$1$:] Provides almost no new diagnostic information.
        \item[$2$:] Carries some informational value but contributes little to narrowing the differential.
        \item[$3$:] Provides some information but is not the most informative option here.
        \item[$4$:] Has clear diagnostic value and meaningfully helps separate candidates.
        \item[$5$:] A pivotal step that markedly reduces diagnostic uncertainty.
    \end{itemize}
    \item \textbf{Cost-effectiveness.} Whether the action is economically and clinically proportionate, considering monetary cost, time, invasiveness, risk, and redundancy with prior examinations.
    \begin{itemize}
        \item[$1$:] Clearly excessive: cost, invasiveness, or risk far outweighs the expected benefit.
        \item[$2$:] Has some necessity but is too costly, burdensome, or complex for the current stage.
        \item[$3$:] Acceptable but not the most economical or appropriate option.
        \item[$4$:] Reasonably priced and appropriately necessary for the current diagnostic stage.
        \item[$5$:] High diagnostic value with low burden and clear necessity.
    \end{itemize}
\end{itemize}

\noindent\textbf{Protocol and Statistics.}
The three raters scored every sampled state independently, with no communication during scoring. Per-state per-method scores are obtained by averaging the three raters' values, and per-method scores reported in the main paper are means over all $300$ sampled states. Inter-rater agreement, measured by Krippendorff's $\alpha$ on the ordinal scale, was above $0.7$ on every dimension, indicating substantial agreement. To test whether CDPR outperforms each baseline, we perform a paired Wilcoxon signed-rank test on the per-state mean scores (CDPR vs.\ baseline) along each of the five dimensions, with Holm correction for multiple comparisons across dimensions; statistical significance is marked in the main result figure.

\end{onecolumn}

\end{document}